\documentclass[runningheads]{llncs}
\usepackage{graphicx}
\usepackage{amssymb}
\usepackage{hyperref}
\usepackage{multirow}  
\usepackage{color}
\usepackage{amsmath,xcolor}
\usepackage{array}
\usepackage[lined,boxed,commentsnumbered,ruled]{algorithm2e}

\begin{document}
\title{Link-aware link prediction over temporal graph by  pattern recognition}

\titlerunning{link-aware link prediction}
%

\author{Bingqing Liu\inst{1,2} \and
Xikun Huang\inst{1,2}}
\authorrunning{Liu et al.}
\institute{Academy of Mathematics and Systems Science, Chinese Academy of Sciences, Beijing 100190, China \and
School of Mathematical Sciences, University of Chinese Academy of Sciences, Beijing 100049, China\\
\email{liubingqing20@mails.ucas.ac.cn}
\email{huangxikun@amss.ac.cn}}

%
%

%

\maketitle              
\begin{abstract}
A temporal graph can be considered as a stream of links, each of which represents an interaction between two nodes at a certain time. On temporal graphs, link prediction is a common task, which aims to answer whether the query link is true or not. To do this task, previous methods usually focus on the learning of representations of the two nodes in the query link. We point out that the learned representation by their models may encode too much information with side effects for link prediction because they have not utilized the information of the query link, i.e., they are link-unaware.  Based on this observation, we propose a link-aware model: historical links and the query link are input together into the following model layers to distinguish whether this input implies a reasonable pattern that ends with the query link.  During this process, we focus on the modeling of link evolution patterns rather than node representations.  Experiments on six datasets show that our model achieves strong performances compared with state-of-the-art baselines, and the results of link prediction are interpretable. The code and datasets are available on the project website: \url{https://github.com/lbq8942/TGACN}.

\end{abstract}

\section{Introduction}
\label{sec:introduction}

Temporal graphs are powerful mathematical abstractions to describe complex dynamic networks and have a wide range of applications in various areas of network science, such as citation networks, communication networks,  social networks, biological networks, and the World Wide Web~\cite{ji2021survey}.  In temporal graph, there are many insightful patterns (usually refers to small and induced temporal subgraph, which is also called motif or graphlet~\cite{sun2019new}) which summarize the evolution laws of links.  To evaluate whether our model captures these patterns, link prediction is a widely used task~\cite{Poursafaei2022TowardsBE}, which is defined in dynamic graphs as, given all historical links before $t$, determine whether link $e=(s,o)$ will happen at timestamp $t$, where $s$ and $o$ are the source and destination nodes respectively. In previous studies, a standard paradigm of link prediction is centered on learning of the temporal node representations of $s$ and $o$, and then the two representations are concatenated and fed into an MLP for link prediction, in which, representation learning is typically operated by the message passing mechanism, i.e., aggregating information from neighborhoods~\cite{Xu2020InductiveRL,Velickovic2018GraphAN,Wang2021APANAP,hamilton2017inductive}.  We note that this is suboptimal for link prediction cause  they do not use the information $e=(s,o)$ when learning temporal representations. In other words, the prior methods are (query) link-unaware for link prediction. 

\begin{figure}
\centering 
\includegraphics[width=0.6\textwidth]{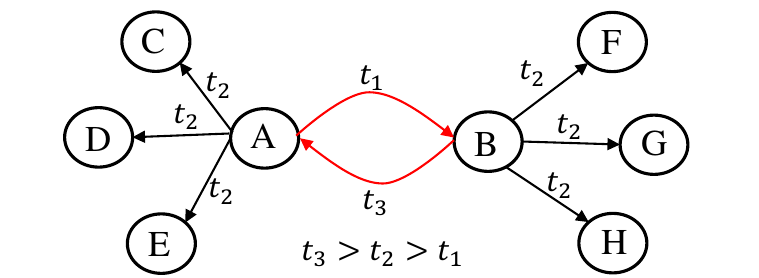}
\caption{An example temporal graph. Temporal subgraph with red links shows an example pattern, i.e., \{(A,B,$t_1$),(B,A,$t_3$)\}, which means that B will  interact with A after A interacts with B. }
\label{fig:exampletg}
\end{figure}
Figure~\ref{fig:exampletg} shows a temporal graph and the temporal subgraph with red links shows an example pattern. The perspective of pattern gives us inspiration for link prediction. For example, the pattern in Figure~\ref{fig:exampletg} tells us that, given the  link (B,A,$t_3$) to be predicted, we only need to look for the existence of the link from A to B in the history (we name this kind of links as target links). Notice that only by first considering (B,A,$t_3$), can we directly check the target links and ignore the other noisy links. However, this is not the case for previous methods. To predict (B,A,$t_3$), previous methods first learn the representations of A and B from historical links. Without taking the query link (B,A,$t_3$) into account, they can easily absorb noisy messages such as (A,C,$t_2$) and (B,F,$t_2$) into their node representations, though we only need the message from (A,B,$t_1$). As a result, much of their aggregated information is not necessary and often has side effects for link prediction because the important information may be diluted by the noisy links. Based on this, we propose a link-aware method, where  historical links and the query link are input together into the following model layers to distinguish whether this input implies a reasonable pattern that ends with the query link. Under the instruction of the query link, our model can directly check the target links. 
\par
\begin{figure}
\centering 
\includegraphics[width=0.8\textwidth]{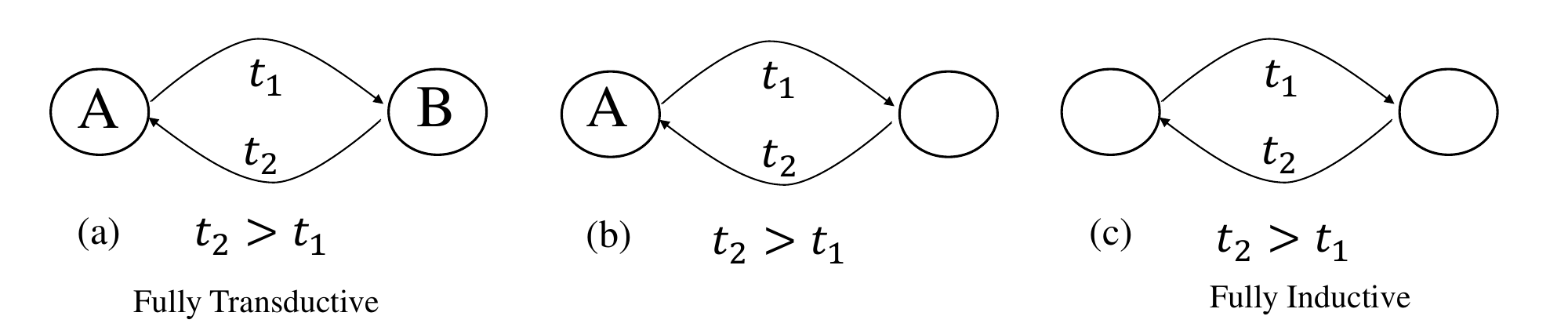}
\caption{Patterns of different granularity.  Pattern (a) shows that B will respond to A after A interacts with B, pattern (b) describes that no matter whom A interacts with, afterwards this person will response to A, whereas pattern (c) reveals that such sequence holds for any two arbitrary nodes.}
\label{fig:pattern_grain}
\end{figure}
Although we have methodological differences for link prediction, for graph learning, we share the following challenges.  \textbf{First, how to make sampling more efficient?} Given the query link, we usually need sample links from the history due to the huge size of the temporal graph. However, previous sampling techniques are usually heuristic methods~\cite{Xu2020InductiveRL,Rossi2020TemporalGN,Wang2021APANAP,Wang2021InductiveRL} and thus not flexible. In order to recall the target links, sometimes they need very long sampling lengths, which significantly increases the number of noisy links.
\textbf{Second, how to learn patterns of different granularity?} Figure~\ref{fig:pattern_grain} shows three patterns of different granularity. Though they are equally important, previous models usually focus only on the extremes, i.e., either fully transductive~\cite{Kumar2019PredictingDE,Trivedi2019DyRepLR,Poursafaei2022TowardsBE} or fully inductive~\cite{Xu2020InductiveRL,Wang2021InductiveRL}. 
\textbf{Third, how to make our results for link prediction interpretable?} Interpretability for link prediction is rarely studied in previous work. APAN~\cite{Wang2021APANAP} has discussed interpretability, however, it  can only tell the importance of historical links to the node representation, which can not be used to interpret the results for link prediction.  In fact, the reason why the previous models are not interpretable can be attributed once again to the fact that they are link-unaware.
\par
To meet the above challenges, here, (1) we develop a link-aware model for link prediction over temporal graphs, which can make full use of the query link when doing link prediction.  (2) we propose a sampling method called parametric sampling to recall historical links that are useful but distant from the query link, which makes our sampling more efficient. (3) To capture patterns of different granularity, two kinds of attention are proposed to perform transductive and inductive learning simultaneously. (4) By equipping our model with technique class activation mapping (CAM)~\cite{Zhou2016LearningDF}, interpretability is easily accessed, which tells us which historical links promote our decision for link prediction. 
\begin{figure*}
\centering 
\includegraphics[width=0.99\textwidth]{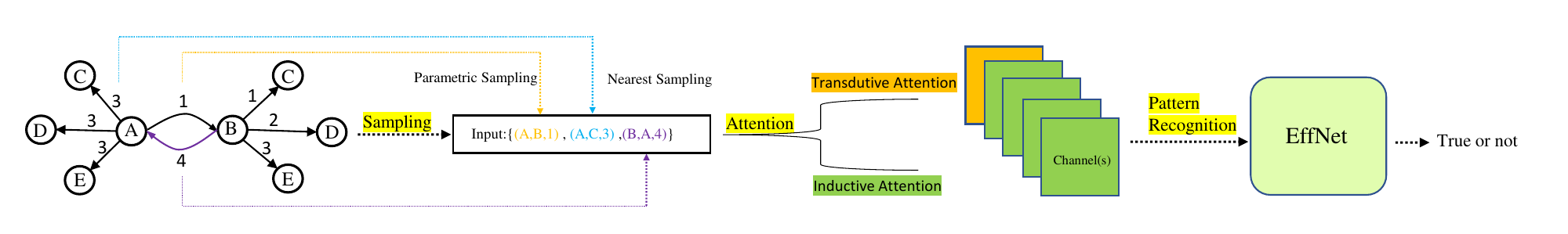}
\caption{The overall pipeline of the proposed Temporal Graph Attention Convolution Network (TGACN). Given the query link (the link from B to A with purple color), TGACN first uses nearest sampling and parametric sampling to recall historical links that may help to predict the query link. Then we encode the input links by two kinds of attention: transductive attention and inductive attention, where node identities information and the inductive structural information are preserved. Finally, a convolutional neural network EffNet is utilized to recognize whether the input implies a reasonable pattern that ends with the query link.}
\label{fig:pipeline}
\end{figure*}
\section{Related Work}
\label{sec:related-work}
Prior work for link prediction over temporal graph mainly focuses on the learning of nodes' representation~\cite{Skarding2021FoundationsAM,chen2020graph}. Early models  are mainly snapshot-based methods, which learns temporal graphs in discrete time space. This kind of method divides the temporal graph into snapshots where links in one snapshot are assumed to take place at the same time and links in different snapshots still maintain the chronological order~\cite{Sankar2020DySATDN,Pareja2020EvolveGCNEG,kazemi2020representation}. Within snapshots, they usually utilize static graph neural networks (GNN)~\cite{kipf2016semi,Velickovic2018GraphAN,hamilton2017inductive} to encode structural features. Between snapshots, they  use RNNs or Transformer~\cite{Vaswani2017AttentionIA} to model the temporal dynamics. The main drawback of these approaches is that they need to predetermine a time granularity for snapshot partition and thus hard to learn structural and temporal dynamics in different time scales. 
 \par
 Learning on temporal graphs in continuous time space recently gained attention. These methods can be broadly classified into two categories, streaming methods and non-streaming methods. Streaming methods maintain a chronologically changing table for representations of all nodes and chronologically predict the links. Each time a new link appears, they use RNNs or Transformer to update that table, i.e., update representations of the source and destination node (and their neighbors) in that new link~\cite{Kumar2019PredictingDE,Trivedi2019DyRepLR,Rossi2020TemporalGN,Wang2021APANAP}. With the updated representations, they can predict the future link.  Given a query link, streaming methods need to first digest all the previous historical links, while non-streaming methods do not~\cite{Wang2021InductiveRL,Xu2020InductiveRL,Wen2022TRENDTE}. It samples only a few historical links and uses GNN or Transformer to aggregate their information to obtain the representations of the two nodes in the query link. Commonly used sampling techniques include nearest sampling~\cite{Rossi2020TemporalGN} and probabilistic sampling~\cite{Wang2021InductiveRL}. 

\section{THE TGACN MODEL}
\label{sec:model}
In this section, we introduce the proposed model, Temporal Graph Attention Convolution Network (TGACN). We first formally give the problem formulation and notations of the link prediction task over a temporal graph, then we introduce our model, which includes three parts: sampling, attention, and pattern recognition.
\subsection{Problem Formulation and Notations}
 A temporal graph can be represented as a stream of links that come in over time, i.e., $E=\{e_1$, $\cdots$,$e_i$,$\cdots\}$ , where link $e_i$  is defined as a triplet $(s_i,  o_i, t_i)$. Link $e_i$ shows that source node $s_i$ interacted with destination node $o_i$  at timestamp $t_i$.   Link prediction requires our model to distinguish the ground truth triplet $(s_g,  o_g, t_g)$ and the corrupted triplet $(s_g,  o_{neg}, t_g)$, where  $o_{neg}$ is sampled from the nodes set $V$. That is, link prediction can be described as: given historical links $e_h=(s_h,o_h,t_h),t_h<t_q$, tell whether query link $e_q=(s_q,o_q,t_q)$ is a ground truth or not. 
 \par
A sequence with $l$ links in chronological order is denoted as $ES=\{e^1,\cdots,e^{l}\}$ and we use $ES(\cdot)\in \mathbb{R}^l$ denotes the corresponding sequence of $\cdot$, for example, $ES(t)=\{t^1$, $ \cdots$, $t^l$\}. Specially, we use $ES(\mathbf{e}) \in \mathbb{R}^{l\times d} $ denotes the vector representations of all links, where boldface $\mathbf{e} \in \mathbb{R}^d$   denotes the vector representation of link $e$. We use matrix $H\in \mathbb{R}^{|V|\times d}$ denotes the trainable representations of all nodes and $H(\cdot) \in \mathbb{R}^d$ denotes the representation of node $\cdot$.
\subsection{Method}
\par
\textbf{Sampling.}
Given a query link, we first need sampling due to the large scale of historical links. Without domain knowledge, a widely used sampling technique is neighborhood sampling~\cite{Velickovic2018GraphAN,Xu2020InductiveRL}, i.e., links that are close to the query link should be sampled with higher priority. However, nearest sampling alone may sometimes fail, for example, in Figure~\ref{fig:pipeline}, when we sample $N=3$ nearest links, the target link (A,B,1) will be missed and  it is not ideal to simply increase the sampling length because it will introduce more  noisy links.  Based on this consideration, for nearest sampling, we still keep $N$ small and propose parametric sampling for those useful but more distant links. Parametric sampling locates valuable historical link  $e_h$ by computing its “closeness" to the query link $e_q$ as follows:
\begin{equation}
\label{eq:attn_1}
\rm{closeness}(e_q,e_h)=\mathbf{e_{q}} \circ \mathbf{e_{h}}
\end{equation}
In which, $\circ$ denotes the dot product between two vectors,  $\mathbf{e_q}$ and $\mathbf{e_h}$ are computed as follows: 
\begin{equation}
\label{eq:eq}
\mathbf{e_q}=\phi(0)+H(s_q)+H(o_q)
\end{equation}
\begin{equation}
\label{eq:eh}
\mathbf{e_h}=\phi(t_q-t_h)+H(s_h)+H(o_h)
\end{equation}
\begin{equation}
\label{eq:timeencoding}
\phi(t)=[cos(\omega_1t+b_1),\cdots,cos(\omega_dt+b_d)]
\end{equation}
Where $\phi(\cdot) \in \mathbb{R}^d$ is a time encoding proposed in TGAT~\cite{Xu2020InductiveRL} and $[\omega_1,b_1, \cdots, \omega_d,b_d] \in \mathbb{R}^{2d}$ are all trainable parameters. 
We calculate the "closeness" of $M$ (usually much larger than $N$) nearest historical links, and select $P$ (usually very small) links with the largest "closeness". Note that despite its potential of recalling useful links that are distant, parametric sampling practically consumes more time than nearest sampling, so we need a trade-off to achieve the best performance.
\par
By nearest sampling and parametric sampling, we sampled a total of $N+P$ historical links, which with the query event together form our input with  $l=(N+P+1)$ links $ES=\{e^1,\cdots,e^l\}$.
\par
\textbf{Attention}.  In order to determine whether there is a reasonable pattern that ends with the query link in the input links, we first need to encode these input links. During the encoding, we should  provide enough convenience for the query link to directly check the target links while preserving as much of the original information as possible. In this paper, the attention mechanism is utilized, specifically includes transductive attention and inductive attention.  Compared with traditional attention mechanism~\cite{Vaswani2017AttentionIA}, we differ in two ways. First, the traditional attention mechanism is based on vector representation and dot-product, while inductive attention is not. Second,  we use only the attention values computed by the "query" and "key", which we believe encode rich pattern information.  In the following, we first introduce the transductive attention, and then the inductive attention. 
\par
\textbf{Transductive Attention}.
Transductive attention tries to extract the pattern information from the representation of input links. Since it is representation-based, the attention values carry rich information of nodes and edges, and thus can be used to capture fine-grained patterns.  We first calculate the vector representations for each link in the input links by eq~\ref{eq:eq} - eq~\ref{eq:timeencoding}. As a result, we get  $ES(\mathbf{e}) \in \mathbb{R}^{l\times d} $. Then we compute the attention values between every two input links using dot product, i.e., we get
	\[
	\begin{bmatrix}
	  \mathbf{e^1} \circ \mathbf{e^1} & \dots&\mathbf{e^1} \circ \mathbf{e^l}\\
	  \vdots&\ddots&\vdots \\
	  \mathbf{e^l} \circ \mathbf{e^1} &\dots&\mathbf{e^l}\circ \mathbf{e^l}
	  \end{bmatrix}_{{l}\times{l}}   
	\]
	We denote the above attention result as $channel(\mathbf{e})\in \mathbb{R}^{l\times l}$
\\

\textbf{Inductive Attention}.
Inductive attention operates directly on node identities and link timestamps rather than vector representations. The goal of this attention is to remove node identities while still preserving the structure of the pattern so that they can be generalized to nodes and links that have not been seen before (see Figure~\ref{fig:pattern_grain}(c)). In order to do this, new attention functions are required.
\par
The timestamps of the input links $ES(t)$  not only reflect the order of occurrence of the links, but the time interval itself carries a wealth of information. To encode this, we use the following attention functions:
\begin{equation}
\label{eq:attn_2}
attn_1(x,y)=\exp(-\alpha|x-y|)
\end{equation}
where time decaying coefficient $\alpha>0$ is a hyperparameter. We first pair the elements in $ES(t)$ and then the above attention function is utilized, i.e., the attention result is:
	\[
	\begin{bmatrix}
	  attn_1(t^1,t^1)&\dots&attn_1(t^1,t^{l})\\
	  \vdots&\ddots&\vdots \\
	  attn_1(t^{l},t^1)&\dots&attn_1(t^{l},t^{l})
	  \end{bmatrix}_{{l}\times{l}}   
	\]
\noindent We denote the above attention result as $channel(t)\in \mathbb{R}^{l\times l}$.
\par
To remove the node identities while still maintain the original pattern topology, we propose to use the following attention function:
\begin{equation}
\label{eq:attn_3}
attn_2(x,y)=\left\{
\begin{array}{rcl}
1 & & {x = y}\\
0 & & else\\
\end{array} \right.
\end{equation}
\par
Like handling $ES(t)$  with eq~\ref{eq:attn_2}, we handle $ES(s)$ and $ES(o)$ with eq~\ref{eq:attn_3}.
Similarly, we can get $channel(s)\in \mathbb{R}^{l\times l}$ and $channel(o)\in \mathbb{R}^{l\times l}$. Note that $ES(s)$ and $ES(o)$ may share the same nodes, while this kind of information is not yet reflected in the above two channels $ES(s)$ and $ES(o)$. To reserve this information, we perform mutual attention between $ES(s)$ and $ES(o)$ and get $channel(s,o)$, i.e.,
	\[
	\begin{bmatrix}
	  attn_2(s^1,o^1)&\dots&attn_2(s^1,o^{l})\\
	  \vdots&\ddots&\vdots \\
	  attn_2(s^{l},o^1)&\dots&attn_2(s^{l},o^{l})
	  \end{bmatrix}_{{l}\times{l}}   
	\]
\par
It is easy to find that the above three channels retain the original pattern structure and do not lose any structural information. That is, given the channels, we can reverse back to the original pattern structure.
\par
The above five channels are finally stacked to form an "image" with shape of $5\times l\times l$, which encodes rich pattern information of these input links, including both nodes identities information and pattern structure information, which helps us capture patterns of different granularity. Note that the above attention functions are all symmetric, without losing any information,  we set half of the "image" being zero.  
\\
\par
\textbf{Pattern Recognition.}
Given the "image", we decide in this part whether it contains a reasonable pattern that ends with the query link. Like in computer vision, a convolutional neural network (CNN) is utilized to carry out this pattern recognition. Here, we directly use the existing CNN architecture, which has already achieved excellent performance in the image classification task.  In this paper, EfficientNetV2-S~\cite{tan2021efficientnetv2} is selected. Since our "image" ( $5\times l\times l$ ) ($l$ is usually less than 20) is much smaller than images ($3 \times 224\times 224$) in the real world, without harming the performance, we further simplify EfficientNetV2-S as EffNet by simply removing the last few stages and tune the number of layers in the left stages. The architecture of EffNet is shown in Table~\ref{table:architecure}.  Besides, to be explainable, a technique called Class Activation Mapping (CAM)~\cite{Zhou2016LearningDF} is utilized, which is able to identify the discriminative regions. For the architecture, CAM requests the utilization of the global average pooling layer before the last classification layer, which is already satisfied by EffNet. 
\par
 To sum up, to be link-aware, TGACN makes use of the query link at two points. First and most importantly, the sampled historical links and the query link are put together as input. Under the instruction of the query link, it is possible for subsequent model layers to directly check the target links. Secondly, we use the vector representation of the query link to conduct parametric sampling, so as to recall historical links that may help to predict the query link.

\section{Experiment}
\subsection{Experimental Setup}
\textbf{Datasets}. We evaluate our method on six widely used datasets, UCI~\cite{Wang2021InductiveRL}, Social Evolution~\cite{Wang2021InductiveRL,Trivedi2019DyRepLR}, Enron~\cite{Sankar2020DySATDN,Wang2021InductiveRL}, Wikipedia~\cite{Kumar2019PredictingDE}, Lastfm~\cite{Kumar2019PredictingDE} and MOOC~\cite{Kumar2019PredictingDE}.   UCI is a network between online posts made by students,  Social Evolution is a network recording the physical proximity between students. Enron is an email communication network. Wikipedia is a network between wiki pages and human editors. Lastfm is a music listening network between users and songs. MOOC is a network of students and online course content units. Table~\ref{table:dataset} briefly shows the statistics of the datasets used in our experiments, more details are referred to~\cite{Poursafaei2022TowardsBE}. For all these datasets, we split them into training, validation, and testing data by chronological order, with the number of links in three datasets scaled to 70\%, 10\%, and 20\% respectively.
\par
\textbf{Baseline Methods}.
  Various kinds of approaches are chosen as benchmark models. DySAT~\cite{Sankar2020DySATDN} and Evolve-GCN~\cite{Pareja2020EvolveGCNEG} are snapshot-based methods. Dyrep~\cite{Trivedi2019DyRepLR}, JODIE~\cite{Kumar2019PredictingDE}, APAN~\cite{Wang2021APANAP} and TGN~\cite{Rossi2020TemporalGN} are streaming methods. For the non-streaming methods, we compared with TGAT~\cite{Xu2020InductiveRL} and CAW-N~\cite{Wang2021InductiveRL}. All of these models had the best performance in their papers. In addtion, we also selected the memory-based method EdgeBank~\cite{Poursafaei2022TowardsBE}, which predicts the query link by simply checking if $(s_q,o_q)$ have ever shown up in the history. Since EdgeBank has not any parameters to learn, thus can help us to check whether our models have learned something. For all baselines, we follow the parameter settings in their papers.

\begin{table}
\caption{The architecture of EffNet, which is simplified from EfficientNetV2-S~\cite{tan2021efficientnetv2}.}
\centering
\scalebox{1.0}{\begin{tabular}{ccccc}
\hline
Stage&\#Layers&Operator&Channel(in/out)\\
\hline
0& 1&Conv(k3*3) & 4/64 \\
1 & 3&Fused-MBConv1(k3*3)&64/64  \\
2 & 7&Fused-MBConv4(k3*3) &64/64 \\
3&1& Conv(k1*1)   &64/1280 \\
4&1& Average Pooling \& FC  &1280/2\\
\hline
\end{tabular}}
\label{table:architecure}
\end{table}  
\begin{table*}
\caption{Statistics of the datasets used in our experiments.}
\centering
\scalebox{1.0}{\begin{tabular}{l|ccccccc}
\hline
&UCI&Social Evo.&Enron&Wikepdia&Lastfm&MOOC\\
\hline
Number of Links&59835&66898&125235&157474&250000&411749\\
Number of Nodes&1899 & 66&184 & 9227&1297&7144\\
\hline
\end{tabular}}
\label{table:dataset}
\end{table*}
\begin{table*}
\caption{AUC performance (in percentage) for link prediction on six datasets. The best results are typeset in bold and the second bests are highlighted with \underline{underline}. All the results are averaged over 10 runs and values in brackets represent the standard deviations.}
\centering
\scalebox{0.87}{\begin{tabular}{c|c|c|c|c|c|c}
	\hline
	Model&UCI&Social Evo.&Enron&Wikipedia&Lastfm&MOOC
	\\
	\hline
	 EdgeBank&81.2(0.02)&63.2(0.03)&82.8(0.03)&94.3(0.02)&83.2(0.04)&47.1(0.10)\\  
	 DyREP&69.6(1.50)&70.6(0.02)&67.2(0.44)&93.2(1.26)&70.1(1.11)&63.8(1.88)\\ 
	 JODIE&82.3(0.75)&85.8(0.12)&85.6(1.35)&94.6(0.60)&80.5(1.09)&89.9(0.28)\\ 
	 DySAT&81.0(0.74)&85.5(0.52)&85.0(0.31)&93.2(0.12)&81.3(1.07)&69.1(1.22)\\
	 EvolveGCN&81.2(0.32)&83.5(0.78)&83.9(0.54)&93.4(0.13)&82.9(1.02)&71.7(1.78)\\
	 TGAT&81.4(0.98)&90.1(0.17)&75.3(0.98)&95.3(0.22)&75.9(1.52)&82.5(1.21)\\ 
	 APAN&90.3(0.72)&83.9(1.20)&83.3(1.62)&98.1(0.09)&85.3(0.52)&88.4(0.45)\\
	 TGN&92.2(0.54)&\underline{92.8(0.43)}&85.5(0.76)&\underline{98.5(0.05)}&83.2(1.47)&\underline{90.6(0.39)}\\ 
	 CAW-N&\underline{94.3(0.41)}&87.2(0.55)&\underline{91.5(0.73)}&\textbf{99.0(0.04)}&\underline{88.0(1.27)}&85.6(0.72)\\
	 \hline
TGACN&\textbf{96.0(0.13)}&\textbf{94.5(0.23)}&\textbf{92.6(0.58)}&\textbf{99.0(0.04)}&\textbf{92.3(0.53)}&\textbf{91.9(0.64)}\\
\hline
\end{tabular}}
\label{table:result}
\end{table*}

\begin{table*}
\caption{Efficiency comparison between TGACN and CAW-N on the convergence speed. The results report the number of epochs required to get the best performance.}
\centering
\scalebox{0.88}{\begin{tabular}{c|c|c|c|c|c|c}
	\hline
	&UCI&Social Evo.&Enron&Wikipedia&Lastfm&MOOC
	\\
	\hline
      CAW-N &\underline{7}&\underline{6}&\underline{16}&\underline{4}&\underline{10}&\underline{20}\\
    TGACN &\textbf{4}&\textbf{3}&\textbf{2}&\textbf{2}&\textbf{4}&\textbf{13}\\
  \hline
\end{tabular}}
\label{table:tg-training-convergence}
\end{table*}

\begin{table*}
\caption{Efficiency comparison between TGACN and CAW-N on training speed. The results report the required training time (minutes per epoch).}
\centering
\scalebox{0.88}{\begin{tabular}{c|c|c|c|c|c|c}
	\hline
	&UCI&Social Evo.&Enron&Wikipedia&Lastfm&MOOC
	\\
	\hline
      CAW-N &\underline{30.3}&\underline{34.9}&\underline{66.7}&\underline{83.7}&\underline{167.3}&\underline{186.3}\\
    TGACN &\textbf{14.0}&\textbf{17.3}&\textbf{66.5}&\textbf{82.5}&\textbf{152.6}&\textbf{77.4}\\
  \hline
\end{tabular}}
\label{table:tg-training-speed}
\end{table*}
\begin{table}
\caption{Ablation study on the sampling methods, two kinds of attention and the architecture of CNN.}
\centering
\scalebox{0.88}{\begin{tabular}{c|c|c|c|c|c|c}
	\hline
	ablation &UCI&Social Evo.& Enron&Wikipedia&Lastfm&MOOC \\  \hline
    remove parametric sampling &96.0&94.5&90.0&98.6&90.3&91.9\\
    remove nearest sampling &94.9&93.6&92.8&98.8&91.3&81.4\\ \hline
    remove transduction attention&96.4&94.4&92.4&99.1&87.9&79.8\\ 
  remove inductive attention &92.0&92.9&90.9&85.7&91.5&77.7\\ 
     \hline
	 use EfficientNetV2-S&95.6&94.6&92.7&98.8&92.2&91.1\\ 
	 use ResNet18&95.8&94.4&92.4&99.0&92.1&90.3\\
	 remove stage 1 and 2 in EffNet&93.6&93.4&92.0&98.9&90.0&86.7\\ 
  \hline
\end{tabular}}
\label{table:ablation}
\end{table}
\begin{table*}
\caption{Hyperparameter investigation on nearest sampling length $N$, parametric sampling length $P$ and time decaying coefficient $\alpha$.}
\centering
\scalebox{0.88}{\begin{tabular}{c|c|c|c|c|c|c}
	\hline
	&UCI&Social Evo.&Enron&Wikipedia&Lastfm&MOOC
	\\
	\hline
      $N$=2 &94.0&92.6&92.4&98.9&91.3&90.7\\
    $N$=12 &96.0&94.5&92.6&99.1&92.3&89.9\\
        $N$=24 &95.4&94.3&91.7&99.1&92.3&92.5\\
  \hline
    	 $P$=0&96.0&94.5&91.7&98.6&89.9&91.9\\ 
	 $P$=6&96.4&92.4&92.2&99.0&92.3&89.5\\
	 $P$=10&96.3&92.4&92.6&99.1&92.2&89.6\\
  \hline
        $\alpha$=1 &95.9&94.4&92.4&98.9&92.2&88.9\\
    $\alpha$=5 &96.0&94.5&92.6&99.0&92.3&91.9\\
        $\alpha$=10 &96.0&94.6&92.3&99.1&92.3&89.2\\
  \hline
\end{tabular}}
\label{table:tg-hp-result}
\end{table*}
\begin{figure}
\centering 
\includegraphics[width=0.70\textwidth]{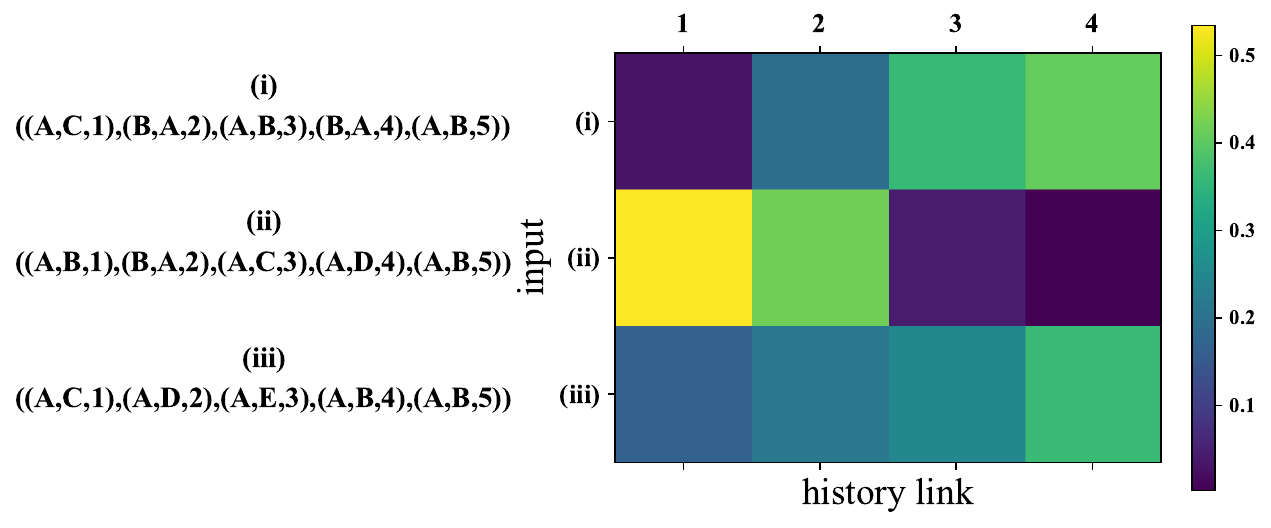}
\caption{Illustration of our model's interpretability. On the left of this Figure, we present three ground truth query links and sample four historical links for each (the end of input is the query link). On the right, by CAM, we show the importance of four historical links to the query link for each input.}
\label{fig:attention}
\end{figure}

\subsection{Results}
In this part, we report the performance of link prediction, which can be seen in Table~\ref{table:result}.  By comparing with the memory-based approach Edgebank, we find that there are some models with worse performance, suggesting that these models are actually not learned well for link prediction (similar conclusion with~\cite{Poursafaei2022TowardsBE}). Results show that we outperform all baselines consistently and significantly on six datasets. We attribute it to that the proposed model is link-aware, thus more advantageous for link prediction. CAW-N, a fully inductive method, generally speaking, achieves the second-best performance. However, CAW-N falls down on some datasets, such as MOOC, a possible reason is that the fully inductive CAW-N can not capture the fine-grained patterns, which harms their performance. In contrast, we consider both transductive learning and inductive learning, which remarkably boosts our performance.  In addition, compared with previous SOTA model CAW-N, TGACN has faster convergence speed and training speed. As presented in  Table~\ref{table:tg-training-convergence}, we usually get the best performance within a few epochs. Table~\ref{table:tg-training-speed} shows that we also have fewer training time per epoch than CAW-N.

\subsection{Ablation Study}
\textbf{Impact of the sampling method.} We explore this by using only the nearest sampling or parametric sampling. As shown in Table~\ref{table:ablation}, generally speaking, these two kind of removals both see a decline in performance, which verifies the effectiveness of our proposed parametric sampling. \textbf{Impact of two kinds of attention.} Table~\ref{table:ablation} shows that remove transductive or inductive attention both brings about a decline in model performance, especially for MOOC, which demonstrates the importance of learning patterns of different granularity. \textbf{Impact of the CNN architecture.}  We first replace EffNet with ResNet18~\cite{he2016deep} and EfficientNetV2-S~\cite{tan2021efficientnetv2} respectively. Table~\ref{table:ablation} shows the robustness of the architecture of CNN, we can see that comparable results are obtained but our EffNet has much fewer parameters. Second, we further simplify EffNet by removing stage 1 and 2, with only a few convolutional layers left. The results show that performance drops but not much. A reasonable conjecture is that the attention values have already encoded  high-level information and thus do not require a very deep neural network to extract. This validates the effectiveness of our proposed attention mechanism for link prediction.

\subsection{Hyperparameter Investigation}
In this part, we investigate how nearest sampling length $N$, parametric sampling length $P$ and time decaying coefficient $\alpha$ in eq~\ref{eq:attn_2} affect the model performance. As shown in Table~\ref{table:tg-hp-result}, generally speaking, model's performance increases with length $N$ and then stabilizes, similar observation can be made for parametric sampling length $P$. Table~\ref{table:tg-hp-result} also shows that the proposed model is robust with respect to time decaying coefficient $\alpha$ for most of the datasets.

\subsection{Interpretability}
In this part, we take the model learned from dataset UCI as an example to conduct a  case study about interpretability. Interpretability requires us to get the impact of historical links on the query link. However, CAM only outputs an "image" with shape of $1\times l\times l$, which marks the importance of each pixel. To translate this importance of pixels to that of links,  we sum the values of $i^{th}$ row and column in that "image" as the importance of the $i^{th}$ historical link to the query link. Figure~\ref{fig:attention} shows three examples for link prediction and illustrates our interpretation for the results of link prediction, i.e., showing the importance of the historical links to the query link. From the heatmap in this Figure, we can see that the query link can directly notice the target links, which verifies the TGACN's effectiveness of being link-aware.

\section{Conclusion}
In this paper, we propose a model named Temporal Graph Attention Convolution Neural Network (TGACN), which is specially designed for link prediction on temporal graphs.  TGACN is as far as we know the first link-aware method  and predict the query link from the perspective of pattern recognition rather than learning the node representations. The empirical results demonstrated that the proposed model has achieved the best performance. We believe that our work provides an alternative, effective way for link prediction.  In the future, we will explore more opportunities for our proposed method. A promising direction is to apply our method to the link prediction of static graph, we leave this for future work.

\bibliographystyle{splncs04}
\bibliography{tgacn}
%




\end{document}